\documentclass{article}
\usepackage{geometry}
\usepackage[utf8]{inputenc} 
\usepackage[T1]{fontenc}    
\usepackage{url}            
\usepackage{booktabs}       
\usepackage{amsfonts}       
\usepackage{nicefrac}       
\usepackage{microtype}      
\usepackage{graphicx}
\usepackage{xcolor}         

\usepackage{comment}
\usepackage{todonotes}

\usepackage{amsthm}
\usepackage{amsmath}
\usepackage{amssymb}
\usepackage{bm}
\usepackage{dsfont}

\newtheorem{theorem}{Theorem}
\newtheorem{lemma}{Lemma}
\newtheorem{proposition}{Proposition}
\theoremstyle{definition}

\makeatletter
\DeclareRobustCommand{\rvdots}{%
  \vbox{
    \baselineskip4\p@\lineskiplimit\z@
    \kern-\p@
    \hbox{.}\hbox{.}\hbox{.}
  }}
\makeatother

\usepackage{graphics}

\definecolor{myred}{RGB}{204, 0, 0}
\definecolor{mygreen}{RGB}{0, 153, 0}
\definecolor{myblue}{RGB}{0, 153, 255}
\definecolor{myorange}{RGB}{255, 153, 51}
\definecolor{mycyan}{RGB}{51, 204, 204}
\definecolor{mypurple}{RGB}{204, 0, 153}

\usepackage{colortbl}
\usepackage{makecell}
\usepackage{multirow}

\newcommand{\best}{\cellcolor{mygreen!50}}

\usepackage[section]{algorithm}
\usepackage{algorithmicx}
\usepackage{algpseudocode}

\newcommand{\minimize}[1]{\operatorname*{minimize}_{#1}\quad}

\newcommand{\st}{\operatorname{subject\ to}\quad}

\newcommand{\Imin}{I_{\rm mb}}
\newcommand{\Iminh}{I_{\rm mb}^{\rm enh}}
\newcommand{\nmin}{n_{\rm mb}}

\newcommand{\II}{\mathds{1}}

\newcommand{\EE}{\mathbb{E}}
\newcommand{\PP}{\mathbb{P}}

\newcommand{\bias}{\operatorname{bias}}

\usepackage{tikz}

\usepackage{pgfplotstable}
\usepgfplotslibrary{groupplots, statistics}

\pgfplotsset{
    compat = newest,
    lineA/.style        = {myblue, very thick},
    lineB/.style        = {myred, very thick, dashed},
    lineC/.style        = {gray, thick, dotted},
    lineD/.style        = {myred, very thick},
    lineDeepTP/.style   = {myblue, very thick},
    lineBaseLine/.style = {gray, very thick, densely dotted},
    lineAPPerf/.style   = {myred, very thick, dash dot},
    linePatMat2/.style  = {myorange, very thick, dash dot},
    linePatMat3/.style  = {mypurple, very thick, dashed},
    lineTFCO/.style     = {mygreen, very thick, dashed},
    scatterpos/.style   = {only marks, myred, fill = myred, fill opacity = 0.15, mark size=1.5pt},
    scatterneg/.style   = {only marks, myblue, fill = myblue, fill opacity = 0.15, mark size=1.5pt},
}

\pgfplotstableread[col sep = comma]{./data/molecules.csv}\MolScoresTopPush
\pgfplotstableread[col sep = comma]{./data/quantile.csv}\quantiles
\pgfplotstableread[col sep = comma]{./data/thresholds.csv}\thresholds
\pgfplotstableread[col sep = comma]{./data/cifar.csv}\AppA
\pgfplotstableread[col sep = comma]{./data/malware.csv}\AppB
\pgfplotstableread[col sep = comma]{./data/scatter.csv}\ExampleScatter

\usetikzlibrary{pgfplots.fillbetween, intersections}
\usetikzlibrary{external}

\usepackage{authblk}

\title{DeepTopPush: Simple and Scalable Method for Accuracy at the Top}
\author[1]{Václav Mácha}
\author[2]{Lukáš Adam}
\author[3]{Václav Šmídl}
\affil[1]{
  Faculty of Nuclear Sciences and Physical Engineering,
  Czech Technical University in Prague,\protect\\
  Prague, Czech Republic
  }
\affil[2]{
  Faculty of Electrical Engineering,
  Czech Technical University in Prague,\protect\\
  Prague, Czech Republic
}
\affil[3]{
  Institute of Information Theory and Automation,
  Czech Academy of Sciences,\protect\\
  Prague, Czech Republic
}

\date{}

\begin{document}

\maketitle

\begin{abstract}
Accuracy at the top is a special class of binary classification problems where the performance is evaluated only on a small number of relevant (top) samples. Applications include information retrieval systems or processes with manual (expensive) postprocessing. This leads to minimizing the number of irrelevant samples above a threshold. We consider classifiers in the form of an arbitrary (deep) network and propose a new method DeepTopPush for minimizing the loss function at the top. Since the threshold depends on all samples, the problem is non-decomposable. We modify the stochastic gradient descent to handle the non-decomposability in an end-to-end training manner and propose a way to estimate the threshold only from values on the current minibatch and one delayed value. We demonstrate the excellent performance of DeepTopPush on visual recognition datasets and two real-world applications. The first one selects a small number of molecules for further drug testing. The second one uses real malware data, where we detected 46\% malware at an extremely low false alarm rate of $10^{-5}$.
\end{abstract}

\section{Introduction}

Binary classifiers compute a score for each sample and compare it with a given threshold to predict whether the sample belongs to the positive or negative class. This score is often interpreted as the probability that the sample is of the positive class, and the threshold is usually $0.5$. Such a task attempts to classify all samples correctly.

On the other hand, many applications need to classify only a small fraction of samples correctly. In information retrieval systems, the user is interested only in the top few queries. Whenever samples undergo manual processing, humans can process only a small fraction of all samples. Cybersecurity defensive mechanisms must have extremely low false positive rates; otherwise, they are removed by the user. In all these applications, the score determines the sample relevance. After computing the scores of all samples, they are compared with a threshold. All samples below the threshold are irrelevant. The small fraction of samples above the threshold is deemed positive. These selected samples are then the search results, items for further analysis, or the detected malware.

Since this task considers only scores above the threshold, \cite{boyd2012accuracy} named it \textit{accuracy at the top}. The important distinction from standard classifiers is that this threshold is no longer fixed, as in the case of $0.5$, but depends on all samples. Therefore, the objective is non-additive and non-decomposable. This brings both theoretical and numerical issues. Standard machine learning algorithms use minibatch sampling. However, when the threshold is computed on a minibatch, it provides a lower estimate of the true threshold. Therefore, the sampled threshold is a biased estimate of the true threshold. Figure~\ref{fig:thresholds1} illustrates this phenomenon. The bias between the true and sampled thresholds is large even for medium-sized minibatches. Backpropagation then propagates this sampling error through the whole gradient, and consequently, the minibatch gradient is a biased estimate of the true gradient. This brings numerical issues \cite{bottou2018optimization}.

\begin{figure}[!ht]
  \centering
  \begin{tikzpicture}
    \begin{axis}[
      height=130,
      width=0.9\linewidth,
      title = {},
      xlabel = {Minibatch size},
      ylabel = {},
      ymin = 1.5,
      ymax = 2.5,
      tick pos = left,
      grid = major,
      grid style = {dashed, draw = gray!50, very thin},
      enlargelimits = false,
      legend pos=south east,
      legend cell align={left},
    ]
      \addplot [lineD]  table[x index=0, y index=1] {\quantiles};
      \addlegendentry{True threshold}
      \addplot [lineA] table[x index=0, y index=2] {\quantiles};
      \addlegendentry{Sampled threshold}
      \addplot [myblue!50,name path=path1] table[x index=0, y index=3] {\quantiles};
      \addplot [myblue!50,name path=path2] table[x index=0, y index=4] {\quantiles};
      \addplot[myblue!50,fill opacity=0.5] fill between[of=path1 and path2]; \addlegendentry{Standard deviation}
    \end{axis}
  \end{tikzpicture}
  \caption{The bias between the sampled and true thresholds computed from scores following the standard normal distribution. The threshold separates the top $1\%$ of samples with the highest scores.}
  \label{fig:thresholds1}
\end{figure}

\begin{figure*}[!ht]
  \centering
  \begin{tikzpicture}
    \begin{axis}[
      width = \textwidth,
      height = 5cm, 
      xmin = -0.525,
      xmax = 2.05,
      ymin = -0.25,
      ymax = 3.25,
      hide axis,
      legend style = {
        column sep = 10pt,
        legend columns = 4,
        legend cell align = {left},
        anchor = south,
        at = {(0.5, 1.05)},
      },
      enlargelimits = false,
      ]
      \addplot [scatterneg] table[x index=1, y index=0] {\ExampleScatter};
      \addlegendentry{Negative scores}
      \addplot [scatterpos] table[x index=3, y index=2] {\ExampleScatter};
      \addlegendentry{Positive scores}
      \addlegendimage{black, very thick, dashed}
      \addlegendentry{Threshold accuracy}    
      \addlegendimage{myorange, very thick, dashed}
      \addlegendentry{Threshold top}    
      \addplot [scatterneg] table[x index=5, y index=4] {\ExampleScatter};
      \addplot [scatterpos] table[x index=7, y index=6] {\ExampleScatter};
      \node [draw, left] at (axis cs:-0.05,2.5) {\parbox{2.2cm}{\textbf{Classifier 1} \\ Accuracy 95\% \\ Top 19\%}};
      \node [draw, left] at (axis cs:-0.05,0.5) {\parbox{2.2cm}{\textbf{Classifier 2} \\ Accuracy 76\% \\ Top 53\%}};
      \draw [black, very thick, dashed] (axis cs:1, 1.75) -- (axis cs:1,3.25);
      \draw [myorange, very thick, dashed] (axis cs:1.764, 1.75) -- (axis cs:1.764,3.25);
      \draw [black, very thick, dashed] (axis cs:0.995,-0.25) -- (axis cs:0.995,1.25);
      \draw [myorange, very thick, dashed] (axis cs:1,-0.25) -- (axis cs:1,1.25);
      \draw [black, very thick] (axis cs:-0.525,1.5) -- (axis cs:2.05,1.5);
      \end{axis}
  \end{tikzpicture}
  \caption{Difference between standard classifiers (top row) and classifiers maximizing accuracy at the top (bottom row). While the former has a good total accuracy, the latter has a good top acccuracy.}
  \label{fig:difference}
\end{figure*}

Our method mitigates this bias. It is based on several results. \cite{li2014top} proposed the TopPush formulation of the accuracy at the top and solved it in its dual formulation. \cite{adam2021general} solved the TopPush formulation directly in its primal form for linear classifiers. Since we generalize the linear TopPush into non-linear classifiers, we name our method \textbf{DeepTopPush}. We stay in the primal form to be able to employ stochastic gradient descent. Due to non-decomposability, we need to propose a way of computing the gradient and reduce the bias mentioned above. Since the threshold always equals to one of the scores \cite{boyd2012accuracy}, its computation has a simple local formula. We implicitly remove some variables and apply the chain rule (backpropagation) to compute the gradient in an end-to-end manner. To reduce the bias, we need to improve the approximation quality of the sampled threshold. We employ again the fact that the true threshold corresponds to one sample. Since this sample changes slowly during optimization, we modify the idea of \cite{adam2019machine} and enhance the current minibatch by the sample, which equalled the sampled threshold on the previous minibatch. As this added sample usually propagates across multiple minibatches, it tracks the threshold, and this trick mitigates the sampled threshold bias. The main contributions of the paper are as follows:
\begin{itemize}\itemsep -0.5mm
  \item We propose \textit{DeepTopPush}, which is a simple and scalable method for accuracy at the top.
  \item We show that \textit{DeepTopPush} increases the computational time only slightly, yet it achieves better performance than prior art methods.
  \item We show both theoretically and numerically that enhancing the minibatch by one sample reduces the bias of the sampled gradient.
\end{itemize}
The paper is organized as follows: Section \ref{sec:theory} introduces a general formulation of accuracy at the top. Section \ref{sec:solving} derives formulas for the bias of the sampled threshold and proposes \textit{DeepTopPush} to minimize it. Section \ref{sec:numerics} shows the good performance of \textit{DeepTopPush} on multiple images recognition datasets, a real-world medical application, and a malware detection dataset, where we detected 46\% malware at an extremely low false alarm rate of $10^{-5}$. To promote reproducibility, our codes are available online.\footnote{\texttt{{\tiny https://github.com/VaclavMacha/AccuracyAtTop.jl}}}

\section{Accuracy at the top}\label{sec:theory}

This section introduces the accuracy at the top. A standard deep network $f$ with weights $\bm w$ takes inputs $\bm x_i$, transforms them into scores $z_i$, and computes the total loss based on these scores and labels $y_i$. On the other hand, accuracy at the top solves
\begin{equation}\label{eq:problem}
  \aligned
  \minimize{w,z,t}&\lambda_1\sum_{i\in I^-}\II_{z_i \ge t} + \lambda_2\sum_{i\in I^+}\II_{z_i < t} \\
  \st &z_i=f(\bm w;\bm x_i), \\
  &t = G(\{(z_i, y_i)\}_{i\in I}).
  \endaligned
\end{equation}
Similarly to the standard network, the classifier $f$ computes the score $z_i$ for each sample $\bm x_i$. Then a general function $G$ takes the scores and labels of \textbf{all} samples and computes the threshold $t$. This makes the problem non-decomposable. The objective function equals the weighted sum of false-positives (negative samples above the threshold) and false-negatives (positive samples below the threshold). Here, $I$, $I^+$ and $I^-$ are the sets of all, positive and negative labels, respectively, and $\II$ is the characteristic ($0/1$) function counting how many times the argument is satisfied. Setting \eqref{eq:problem} includes TopPush \cite{li2014top} which minimizes the number of positive samples below the highest-ranked negative sample. This fits into \eqref{eq:problem} with $\lambda_1=0$, $\lambda_2=1$ and $t=\max_{i\in I^-} z_i$.

Figure \ref{fig:difference} shows the difference between the standard approach with cross-entropy and accuracy at the top. While classifier 1 has good total accuracy, its top accuracy is subpar because of the few negative outliers. On the other hand, classifier 2 has worse total accuracy, but its top accuracy is extremely good because more than half of the positive samples are on the top. While classifier 1 selected different thresholds for the accuracy and top metrics, these thresholds coincide for classifier 2.

Table \ref{table:summary} shows other special cases of \eqref{eq:problem} including maximizing precision at a given level of recall \cite{mackey2018constrained} or recall at $K$. The threshold $t$ always equals to the sample with the $j^*$-th highest score on all, positive, or negative samples. The problems differ only in $j^*$ and from which samples the threshold is computed. For example, Pat\&Mat-NP \cite{adam2021general} minimizes the false negative rate (equivalently maximizes the true positive rate) under the constraint that the false positive rate is at most $\tau$.

\begin{table}[!ht]
  \centering
  \begin{tabular}{@{}llllll@{}}
    \toprule
    Name & $\lambda_1$ & $\lambda_2$ & $t$ computed from & $j^*$ \\
    \midrule
    Prec@Rec    & $1$ & $0$ & positive samples & $n^+\tau$ \\
    Rec@K       & $0$ & $1$ & all samples      & $K$ \\
    TopPush     & $0$ & $1$ & negative samples & $1$ \\
    Pat\&Mat-NP & $0$ & $1$ & negative samples & $n^-\tau$ \\
    \bottomrule
  \end{tabular}
  \caption{Selected problems of setting \eqref{eq:problem}. False-positives and false-negatives have weights $\lambda_1$ and $\lambda_2$, the threshold $t$ equals to the sample with the $j^*$-th highest score on all, positive, or negative samples.}
  \label{table:summary}
\end{table}

\subsection{Related works}

There is a close connection between accuracy at the top and ranking problems \cite{batmaz2019review,werner2019review}. This was, together with similarities to the Neyman-Pearson problem, showed in \cite{adam2021general}. A special case of the ranking problems attempts to rank positive samples above negative samples. Several approaches, such as RankBoost~\cite{freund2003efficient}, Infinite Push~\cite{agarwal2011infinite} or $p$-norm push~\cite{rudin2009p} employ a positive-negative pairwise comparison of scores, which can handle only small datasets. TopPush~\cite{li2014top} converts the pairwise sum into a single sum and minimizes the false-negatives below a threshold given by the maximum score corresponding to negative samples. Thus, it converts ranking into accuracy at the top problems.

Two approaches for solving \eqref{eq:problem} exist. The first approach considers the threshold constraint as it is, while the second approach uses heuristics to approximate it. In the first approach, Acc@Top~\cite{boyd2012accuracy} argues that the threshold equals one of the scores. They fix the index of a sample and solve as many optimization problems as there are samples. \cite{eban2016scalable,adam2021general,kumar2021implicit} write the threshold as a constraint and replace both the objective and the constraint via surrogates. \cite{eban2016scalable} uses Lagrange multipliers to obtain a minimax problem, \cite{mackey2018constrained} implicitly removes the threshold as an optimization variable and uses the chain rule to compute the gradient while \cite{macha.adam.smidl.2020} solves an SVM-like dual formulation with kernels. \cite{grill2016learning} uses the same formulation but applies surrogates only to the objective and recomputes the threshold after each gradient step. TFCO \cite{cotter2019optimization} solves a general class of constrained problems via a minimax reformulation. In the second approach, SoDeep \cite{engilberge2019sodeep} or SmoothI \cite{thonet2021smoothi} use the fact that the threshold may be easily computed from sorted scores. They approximate the sorting operator by a network trained on artificial data. AP-Perf \cite{fathony2019ap} considers a general metric and hedges against the worst-case perturbation of scores. The authors argue that the problem is bilinear in scores and use duality arguments. However, the bilinearity is lost when optimizing with respect to the weights of the original network. 

\section{DeepTopPush as a method for maximizing accuracy at the top}\label{sec:solving}

This section first shows a basic algorithm to solve \eqref{eq:problem}. We then argue that the stochastic gradient descent produces a biased estimate of the true gradient, and we mention two strategies for mitigating this bias. Based on one strategy, we propose the \textit{DeepTopPush} algorithm. The whole section assumes that the classifier $f$ is differentiable.

\subsection{Basic algorithm for solving accuracy at the top}

Even though the presented technique can be applied to any formulation from Table \ref{table:summary}, for simplicity, we derive it only for the TopPush formulation, where $\lambda_1=0$ and $\lambda_2=1$. This amounts to minimizing the false-negatives in \eqref{eq:problem}. Since the function $\II$ in the formulation \eqref{eq:problem} is discontinuous, it is usually replaced by a general surrogate function $l$ which is continuous and non-decreasing. This leads to
\begin{equation}\label{eq:problem_surr1}
  \aligned
  \minimize{w,z,t}& \frac{1}{n^+}\sum_{i\in I^+}l(t-z_i) \\
  \st&z_i=f(\bm w;\bm x_i), \\
  &t = G(\{(z_i, y_i)\}_{i\in I}).
  \endaligned
\end{equation}
To apply the stochastic gradient descent, we need to compute the gradient. The core idea follows \cite{mackey2018constrained} which was proposed in a more general context in \cite{adam2019machine}. It rewrites problem \eqref{eq:problem_surr1} into its equivalent form
\begin{equation}\label{eq:problem_surr2}
  \minimize{w} \frac{1}{n^+}\sum_{i\in I^+}l\big(G\big(\{(f(\bm w;\bm x_j), y_j)\}_{j\in I}\big) - f(\bm w;\bm x_i)\big).
\end{equation}
This form removed the constraints and it has the advantage that the only optimization variable is $\bm w$ instead of $(\bm w, \bm z, t)$ in \eqref{eq:problem_surr1}. In all cases from Table \ref{table:summary}, the threshold $t$ always equals to one of the scores, let it have index $j^*$ and then $t=z_{j^*}$. Denoting the objective of \eqref{eq:problem_surr2} by $L(\bm w)$, the chain rule implies that the gradient of the objective from \eqref{eq:problem_surr2} equals to
\begin{equation}\label{eq:grad1}
  \nabla L(\bm w) = \frac{1}{n^+} \sum_{i\in I^+}l'(t-z_i)\big(\nabla_w f(\bm w;\bm x_{j^*}) - \nabla_w f(\bm w;\bm x_i)\big).
\end{equation}
The stochastic gradient descent replaces the sum over all positive samples $I^+$ with a sum over all positive samples in a minibatch $\Imin^+$. However, as both the threshold $t$ and the index $j^*$ depend on all scores $z_i$, they need to be approximated on the minibatch as well. We denote these approximations by $\hat t$ and $\hat j$, respectively. Denoting the number of positive samples in the minibatch by $\nmin^+$, we replace the true gradient \eqref{eq:grad1} by the \textit{sampled gradient}
\begin{equation}\label{eq:grad2}
  \nabla \hat L = \frac{1}{\nmin^+}\sum_{i\in \Imin^+}l'(\hat t-z_i)\big(\nabla_w f(\bm w;\bm x_{\hat j}) - \nabla_w f(\bm w;\bm x_i) \big),
\end{equation}
The most straightforward way is to choose the sampled threshold $\hat t$ by the same rule as the true threshold $t$. As an example, if $t$ is the $100^{\rm th}$ largest score on the whole dataset and $\frac{n}{\nmin}=20$ is the ratio of sizes of the whole dataset and of the minibatch, we select the sampled threshold $\hat t$ as the $5^{\rm th}$ largest score on the minibatch. We summarize this procedure in Algorithm \ref{alg1}.

\begin{figure*}
  \begin{minipage}{0.48\textwidth}
    \begin{algorithm}[H]
      \centering
      \begin{algorithmic}[1]
        \State Initialize weights $\bm w$
        \Repeat
        \State Select minibatch $\Imin$
        \State \phantom{$\Iminh$}
        \State Compute $z_i\gets f(\bm w;\bm x_i)$ for $i\in\Imin$
        \State Set $\hat t \gets G(\{(z_i,y_i)\}_{i\in\Imin})$
        \State 
        \State Compute $\nabla \hat L$ based on $\Imin$\phantom{$\Iminh$}
        \State Make a gradient step
        \Until{stopping criterion is satisfied}
      \end{algorithmic}
      \caption{Basic algorithm for solving \eqref{eq:problem} \\}
      \label{alg1}
    \end{algorithm}
  \end{minipage}
  \hfill
  \begin{minipage}{0.48\textwidth}
    \begin{algorithm}[H]
      \centering
      \begin{algorithmic}[1]
        \State Initialize weights $\bm w$, random index $j^*$
        \Repeat
        \State Select minibatch $\Imin$
        \State Enhance minibatch $\Iminh = \Imin \cup \{j^*\}$
        \State Compute $z_i\gets f(\bm w;\bm x_i)$ for $i\in\Iminh$
        \State Set $\hat t \gets \{\max z_i \mid i\in \Iminh \cap I^-\}$
        \State Find index $j^*$ such that $t = z_{j^*}$
        \State Compute $\nabla \hat L$ based on $\Iminh\cap I^+$
        \State Make a gradient step
        \Until{stopping criterion is satisfied}
      \end{algorithmic}
      \caption{DeepTopPush as an efficient method for maximizing accuracy at the top.}
      \label{alg2}
    \end{algorithm}
  \end{minipage}
\end{figure*}

\subsection{Bias of the sampled gradient}

Convergence proofs of the stochastic gradient descent require that the sampled gradient is an unbiased estimate of the true gradient \cite{bottou2018optimization}. This means that
\begin{equation}\label{eq:defin_bias}
  \bias(\bm w) := \nabla L(\bm w) - \EE \nabla \hat L(\bm w)
\end{equation}
equals to $0$ for all $\bm w$. A comparison of \eqref{eq:grad1} and \eqref{eq:grad2} shows that a necessary condition is that the sampled threshold $\hat t$ is an unbiased estimate of the true threshold $t$. However, the sampled version underestimates the true value, which is evident for the maximum where the sampled maximum is always smaller or equal to the true maximum. The next result quantifies the difference between the sampled and true thresholds.

\begin{proposition}[\cite{Glynn96importancesampling}]\label{proposition:bound}
  Let $X$ be an absolutely continuous random variable with distribution function $F$, let $X_1,\dots,X_n$ be iid samples from $X$ and let $\tau\in(0,1)$. Denote the true threshold $t=F^{-1}(1-\tau)$ and the sampled threshold $\hat t=X_{[\lceil n\tau\rceil]}$. If $F$ is differentiable with a positive gradient at $t$, then
  \begin{equation*}
    \sqrt{n}(t - \hat t) \rightarrow N\left(0, \frac{\tau(1-\tau)}{F'(t)^2}\right),
  \end{equation*}
  where the convergence is in distribution and $N$ denotes the normal distribution.
\end{proposition}

This proposition states that when the minibatch size increases to infinity, the variance of the sampled threshold is approximately $\frac{\tau(1-\tau)}{nF'(t)^2}$. Figure \ref{fig:thresholds1} in the introduction shows this empirically for the case where the scores follow the standard normal distribution and $\tau=0.01$ is the desired top fraction. The approximation is poor with both large bias and standard deviation. Even though this result gives us insight into the bias of the sampled threshold, we are ultimately interested in the bias of the sampled gradient $\nabla \hat L(\bm w)$. To do so, recall that $j^*$ is the threshold index on the whole dataset ($t=z_{j^*}$) while $\hat j$ is the threshold index on the minibatch ($\hat t=z_{\hat j}$). We split the computation based on whether these two indices are identical.

\begin{lemma}\label{lemma:convergence}
  Let $j^*$ be unique. Assume that the selection of positive and negative samples into the minibatch is independent and that the threshold is computed from negative samples while the objective is computed from positive samples. Then the conditional expectation of the sampled gradient satisfies
  \begin{equation*}
    \EE\left(\nabla \hat L(\bm w) \mid \hat j=j^*\right ) =  \nabla L(\bm w).
  \end{equation*}
\end{lemma}
\begin{proof}
  If $j^*$ is unique, then the true threshold $t$ is a differentiable function. The differentiability of $L$ and $\hat L$ follows from the chain rule. If $\hat j=j^*$ holds, then the sampled gradient equals to
  \begin{equation}\label{eq:grad_min_aux}
    \nabla \hat L(\bm w)= \frac{1}{\nmin^+}\sum_{i\in \Imin^+}l'(t-z_i)\big(\nabla_w f(\bm w;\bm x_{j^*}) - \nabla_w f(\bm w;\bm x_i) \big).
  \end{equation}
  The summands are identical to the ones in \eqref{eq:grad1}. Since the sum is performed with respect to positive samples, the threshold is computed from negative samples, the lemma statement follows.
\end{proof}

Now we present the main result about the bias.

\begin{theorem}\label{theorem:convergence}
  Under the assumptions of Lemma \ref{lemma:convergence}, the bias of the sampled gradient from \eqref{eq:defin_bias} satisfies
  \begin{equation}\label{eq:comp_bias}
    \bias(\bm w) = \PP(\hat j\neq j^*) \left(\nabla L(\bm w) - \EE\left(\nabla \hat L(\bm w) \mid \hat j\neq j^*\right) \right).
  \end{equation}
\end{theorem}
\begin{proof}
  The law of total expectation implies
  $$
    \aligned
    \EE \nabla \hat L(\bm w) &= \PP(\hat j=j^*)\EE(\nabla \hat L(\bm w) \mid \hat j=j^*) \\
    &\qquad + \PP(\hat j\neq j^*)\EE(\nabla \hat L(\bm w) \mid \hat j\neq j^*),
    \endaligned
  $$
  from where the statement follows due to definiton \eqref{eq:defin_bias} and Lemma \ref{lemma:convergence}.
\end{proof}

The assumptions of Theorem \ref{theorem:convergence} holds for all methods from Table \ref{table:summary} with the exception of Rec@K. For this method, the bias contains an additional term, as we show in the appendix.

The bias \eqref{eq:comp_bias} consists of a multiplication of two terms. We propose two strategies for reducing the bias. The first strategy reduces both terms, while the second strategy reduces only the first term.

\subsection{Bias reduction: Increasing minibatches size}\label{sec:bias1}

The natural choice to mitigate the bias is to work with large minibatches. Even though this is not a standard way, some works suggest this route \cite{you2019large}. When the minibatch is large, it contains more samples and the chance that $\hat j$ differs from $j^*$ decreases. This reduces the first term in~\eqref{eq:comp_bias}. Moreover, Proposition \ref{proposition:bound} ensures that the difference between the sampled threshold $\hat t$ and the true threshold $t$ is small. Then the difference between the true gradient \eqref{eq:grad1} and the sampled gradient \eqref{eq:grad2} decreases as well. This reduces the second term in \eqref{eq:comp_bias}. This approach is applicable to any method from Table~\ref{table:summary}.

\subsection{Bias reduction: Incorporating delayed values}\label{sec:bias2}

Various reasons may enforce the use of small minibatches. Then Algorithm \ref{alg1} is not suitable for a small fraction of top samples. For example, a minibatch of size $32$ with $16$ negative samples must have thresholds $\tau\ge \frac{100}{16}=6.25\%$. However, we need to aim for much smaller thresholds.

We propose a simple fix based on the reasoning that when the weights $\bm w$ of a neural network are updated, the scores $\bm z$ usually do not change much, especially for a small learning rate. This means that if a sample has the largest score, it will likely have the largest score even after the gradient step. Since the threshold $t$ for TopPush equals the largest score corresponding to negative samples, we can easily track it. We enhance the current minibatch by the negative sample from the previous minibatch with the highest score. This significantly increases the chance that the sampled threshold is the true threshold and, due to the first term in \eqref{eq:comp_bias}, reduces the bias of the sampled gradient.

We summarize the procedure in Algorithm \ref{alg2} and show it next to Algorithm \ref{alg1} to highlight the differences. In every iteration, it stores the index $j^*$ of the sample, which equals the threshold (step~7). We add it to the enhanced minibatch (step 4). Since we can track only the maximum, we set the threshold as the maximum of scores from negative samples (step~6) and minimize false-positives. Since Algorithm \ref{alg2} uses the same formulation as \textit{TopPush} \cite{li2014top} but can handle an arbitrary classifier, we name it \textit{DeepTopPush}. We provide empirical evidence of why our technique works later in Section \ref{sec:delay}.

\section{Numerical experiments}\label{sec:numerics}

This section presents numerical results for \textit{DeepTopPush}. Table \ref{table:summary} shows that it is similar to \textit{Pat\&Mat-NP}. While the former maximizes the number of positives above the largest negative, while the latter maximizes the number of positives above the $n^-\tau$-largest negative. The former may be understood as requiring no false-positives, while the latter allows for false positive rate $\tau$.

Section \ref{sec:bias1} showed that we can use large minibatches to obtain good results for \textit{Pat\&Mat-NP} for small fractions of top samples $\tau$. Section~\ref{sec:bias2} showed that \textit{DeepTopPush} works well even with small minibatches if we track the threshold by enhancing the minibatch by one sample. We present numerical comparisons in several sections, each with a different purpose. Comparison with the prior art \textit{TFCO} and \textit{AP-Perf} is performed on several visual recognition datasets and shows that \textit{DeepTopPush} outperforms other methods. Then we present two real-world applications. The first one shows that \textit{DeepTopPush} can handle ranking problems. The second one presents results on a complex malware detection problem. Finally, we show similarities between \textit{DeepTopPush} and \textit{Pat\&Mat-NP} and explain why enhancing the minibatch in Algorithm \ref{alg2} works.

\subsection{Dataset description and Computational setting}\label{sec:set}

We consider the following image recognition datasets: FashionMNIST~\cite{xiao2017fashion}, CIFAR100~\cite{krizhevsky2009learning}, SVHN2~\cite{netzer2011reading} and ImageNet~\cite{russakovsky2015imagenet}. These datasets were converted to binary classification tasks by selecting one class as the positive class and the rest as the negative class. ImageNet merged turtles and non-turtles. We also consider the 3A4 dataset~\cite{ma2015deep} with molecules and their activity levels. Finally, malware analysis reports of executable files were provided by a cybersecurity company. This is an extremely tough dataset as individual samples are JSON files whose size ranges from 1kB to 2.5MB. Moreover, they contain different features, and their features may have variable lengths. All datasets are summarized in the appendix.

We use truncated quadratic loss~$l(z) = (\max\{0, 1 + z\})^2$ as the surrogate function and $\tau=\frac{1}{n^-}$ and $\tau=0.01$. This first one computes the true positive rate above the second highest-ranked negative, while the latter allows for the false positive rate of $1\%$. All algorithms were run for $200$ epochs on an NVIDIA P100 GPU card with balanced minibatches of 32 samples. The only exception was Malware Detection, which was run on a cluster in a distributed manner, and where the minibatch size was $20000$. For the evaluation of numerical experiments, we use the standard receiver operating characteristic (ROC) curve. All results are computed from the test set. All codes were implemented in the Julia language~\cite{bezanson2017julia}. The network structure was the same for all methods; we describe them in the online appendix.

\subsection{Comparison with prior art}\label{sec:comparison}

We compare our methods with \textit{BaseLine}, which uses the weighted cross-entropy. Moreover, we use two prior art methods which have codes available online, namely \textit{TFCO} \cite{cotter2019optimization,narasimhan2019optimizing} and \textit{AP-Perf} \cite{fathony2019ap}. We did not implement the original TopPush because its duality arguments restrict the classifiers to only linear ones. Table \ref{table:time} shows the time requirement per epoch. All methods besides \textit{AP-Perf} have similar time requirements, while \textit{AP-Perf} is much slower. This difference increases drastically when the minibatch size increases, as noted in \cite{fathony2019ap}. We do not present the results for SVHN for \textit{AP-Perf} because it was too slow and for \textit{TFCO} because we encountered a TensorFlow memory error. All these methods are designed to maximize true-positives when the false positive rate is at most $\tau$. This is the same as for \textit{Pat\&Mat-NP}.

\begin{table}[!ht]
  \centering
  \begin{tabular}{@{}llll@{}}
      \toprule      
       & FashionMNIST & CIFAR100 & SVHN \\ \midrule
      BaseLine & 4.4s & 5.1s & 62.8s \\
      DeepTopPush & 4.8s & 5.6s & 66.6s \\
      Pat\&Mat-NP & 4.8s & 5.6s & 66.6s \\
      TFCO & 7.2s & 6.5s & - \\
      AP-Perf & 95.3s & 81.2s & - \\
      \bottomrule
  \end{tabular}
  \caption{Time requirements per epoch for investigated methods for minibatches of size $\nmin=32$.}
  \label{table:time}
\end{table}

\begin{table*}[ht]
  \centering
  \footnotesize
  \begin{tabular}{@{}c|llllll@{}}
      \toprule
      & \thead{Dataset} & \thead{BaseLine}
                        & \thead{DeepTopPush}
                        & \thead{Pat\&Mat-NP}
                        & \thead{TFCO}
                        & \thead{AP-Perf} \\
      \midrule
      \multirow{4}{*}{\rotatebox[origin=c]{90}{\parbox[c]{1.5cm}{\centering tpr@fpr $\tau=\nicefrac{1}{n^-}$}}}
      & FashionMNIST & $5.06 \pm 1.41$
                     & \best $27.30 \pm 5.91$
                     & $22.21 \pm 5.62$
                     & $11.30 \pm 3.44$
                     & $9.90$ \\
      & CIFAR100   & $1.70 \pm 0.46$
                   & \best $14.40 \pm 5.44$
                   & $8.10 \pm 3.45$
                   & $7.70 \pm 2.28$
                   & $5.00$ \\
      & 3A4        & $2.58 \pm 0.61$ 
                   & \best $5.61 \pm 1.70$
                   & $3.79 \pm 0.90$
                   & $3.03 \pm 1.52$
                   & $3.03$ \\
      & SVHN       & $6.51 \pm 1.37$
                   & \best $12.21 \pm 5.39$
                   & $12.07 \pm 4.41$ 
                   & - & -\\
      \midrule
      \multirow{4}{*}{\rotatebox[origin=c]{90}{\parbox[c]{1.5cm}{\centering tpr@fpr $\tau=0.01$}}}
      & FashionMNIST & $63.14 \pm 1.39$
                     & \best $75.37 \pm 1.18$
                     & $74.11 \pm 1.00$
                     & $73.27 \pm 2.92$
                     & $64.60$ \\
      & CIFAR100   & $49.40 \pm 4.90$
                   & \best $70.20 \pm 2.14$
                   & $66.30 \pm 2.33$
                   & $67.30 \pm 1.79$
                   & $65.00$ \\
      & 3A4        & $57.80 \pm 0.35$ 
                   & $60.08 \pm 3.35$
                   & \best $65.91 \pm 0.59$
                   & $54.55 \pm 10.22$
                   & $63.64$ \\
      & SVHN       & $84.72 \pm 0.84$
                   & $91.05 \pm 1.45$
                   & \best $91.07 \pm 0.30$
                   & - & - \\
      \bottomrule
  \end{tabular}
  \caption{The true positive rates (in \%) at two levels of false positive rates averaged across ten indepenedent runs with standard deviation. The best methods are highlighted.}
  \label{tab:Overall comparison}
\end{table*}

Table \ref{tab:Overall comparison} shows the true positive rate (tpr) above the second-largest negative and at the prescribed false positive rate (fpr) $\tau=0.01$. Using the second-largest negative, which corresponds to $\tau=\frac{1}{n^-}$, allows for one outlier. The results are averaged over ten independent runs except for AP-Perf, which is too slow. The best result for each metric (in columns) is highlighted. All methods are better than \textit{BaseLine}. This is not surprising as all these methods are designed to work well for low false positive rates. \textit{DeepTopPush} outperforms all other methods at the top, while it performs well at the low fpr of $\tau=0.01$. There \textit{Pat\&Mat-NP}, which also falls into our framework, performs well. Both these methods outperform the state of the art methods. 

Figure \ref{fig: roc curves} \textbf{A)} shows the ROC curves on CIFAR100 averaged over ten independent runs. We use the logarithmic $x$ axis to highlight low fpr modes. \textit{DeepTopPush} performs significantly the best again whenever the false positive rate is smaller than $0.01$.

As a further test, we performed a simple experiment on ImageNet. We modified the pre-trained EfficientNet B0 \cite{tan2019efficientnet} by removing the last dense layer and adding another dense layer with one output. Then we retrained the newly added layer to perform well at the top. The original EfficientNet achieved $68.0\%$ at the top, while \textit{DeepTopPush} achieved $70.0\%$ for the same metric. This shows that \textit{DeepTopPush} can provide better accuracy at the top than pre-trained networks.

\subsection{Application to ranking}

The 3A4 dataset contains information about activity levels of approximately $50000$ molecules, each with about $10000$ descriptors. The activity level corresponds to the usefulness of the molecule for creating new drugs. Since medical scientists can focus on properly investigating only a small number of molecules, it is important to select a small number of molecules with high activity.

We converted the continuous activity level into binary by considering a threshold on the activity. Since the input is large-dimensional, and there is no spatial structure to use convolutional neural networks, we used PCA to reduce the dimension to $100$. Then we created a network with two hidden layers and applied \textit{DeepTopPush} to it. The test activity was evaluated at the continuous (and not binary level). Table \ref{tab:Overall comparison} shows again the results at the top. \textit{DeepTopPush} outperforms other methods. Figure \ref{fig:molecules} shows that high scores (output of the network) indeed correspond to high activity. Thus, even though the problem was ``binarized'' and its dimension reduced, our algorithm was able to select a small number of molecules with high activity levels. These molecules can be used for further manual (expensive) investigation.

\begin{figure}[!ht]
  \centering
  \begin{tikzpicture}
    \begin{axis}[
      width=0.5\linewidth,
      title = {},
      xlabel = {Scores},
      ylabel = {Activity},
      xmax = 8,
      ymax = 8.5,
      tick pos = left,
      grid = major,
      grid style = {dashed, draw = gray!50, very thin},
      enlargelimits = false,
      ]
      \addplot [scatterneg] table[x index=0, y index=1] {\MolScoresTopPush};
      \addplot [scatterpos] table[x index=2, y index=3] {\MolScoresTopPush};
      \node[circle, black, draw, very thick, minimum size = 30pt, label = above left:Target] at (axis cs:5.2,7.55) {};
      \end{axis}
  \end{tikzpicture}
  \caption{Results for the 3A4 dataset. The goal was to assign large scores to a few molecules with high activity (scores on top-right are preferred).}
  \label{fig:molecules}
\end{figure}

\begin{figure*}[!ht]
  \centering
  \begin{tikzpicture}
    \begin{groupplot}[
        group style = {
            group size = 2 by 1,
            horizontal sep = 20pt,
            x descriptions at = edge bottom,
            y descriptions at = edge left,
        },
        footnotesize,
        width= 0.4\linewidth,
        xlabel = {False positive rate},
        ylabel = {True positive rate},
        ytick = {0, 0.2, 0.4, 0.6, 0.8, 1},
        ymin = -0.05,
        xmode = log,
        tick pos = left,
        grid = major,
        grid style = {dashed, draw = gray!50, very thin},
        enlargelimits = false,
        height = 150,
    ]

    \nextgroupplot[
      title = {\textbf{A)} CIFAR100},
      xmin = 0.001,
      legend to name={CommonLegend},
      legend style = {
          column sep = 10pt,
          legend columns = 3,
          legend cell align = {left},
      }
    ]
      \addlegendimage{lineDeepTP}
      \addlegendentry{DeepTopPush}
      \addlegendimage{lineBaseLine}
      \addlegendentry{BaseLine}
      \addlegendimage{lineTFCO}
      \addlegendentry{TFCO($\tau = 10^{-2}$)}
      \addlegendimage{lineAPPerf}
      \addlegendentry{AP-Perf($\tau = 10^{-2}$)}
      \addlegendimage{linePatMat3}
      \addlegendentry{Pat\&Mat-NP($\tau = 10^{-3}$)}
      \addlegendimage{linePatMat2}
      \addlegendentry{Pat\&Mat-NP($\tau = 10^{-2}$)}

      \addplot [lineDeepTP] table[x index=2, y index=3] {\AppA};
      \addplot [lineTFCO] table[x index=12, y index=13] {\AppA};
      \addplot [lineBaseLine] table[x index=0, y index=1] {\AppA};
      \addplot [lineAPPerf] table[x index=8, y index=9] {\AppA};
      \draw (axis cs:0.01,-0.05) -- (axis cs:0.01,1) node [right] at (axis cs:0.01,0.3) {Optimized threshold};

    \nextgroupplot[
      title = {\textbf{B)} Malware detection},
      xmin = 0.00001,
    ]
      \addplot [lineDeepTP]  table[x index=4, y index=5] {\AppB};
      \addplot [linePatMat3]  table[x index=6, y index=7] {\AppB};
      \addplot [linePatMat2] table[x index=0, y index=1] {\AppB};
      \addplot [lineBaseLine]  table[x index=2, y index=3] {\AppB};
      \fill [myblue] (axis cs:0.000015,0.459232673) circle [radius=4pt];
      \fill [mypurple] (axis cs:0.001,0.924199612) circle [radius=4pt];
      \fill [myorange] (axis cs:0.01,0.915815306) circle [radius=4pt];

    \end{groupplot}
    \path ($(group c1r1.north)!.5!(group c2r1.north)$) ++ (0,1.1) node {\pgfplotslegendfromname{CommonLegend}};
  \end{tikzpicture}
  \caption{\textbf{A)} ROC curves averaged over ten runs on the CIFAR100 dataset. \textbf{B)} ROC curve for Malware Detection dataset. The circles show the thresholds the methods were optimized for.}
  \label{fig: roc curves}
\end{figure*}

\subsection{Real-world application}

This section shows a real-world application of the accuracy at the top. A renowned cybersecurity company provided malware analysis reports of executable files. Its structure is highly complicated because each sample has a different number of features, and features may have a complicated structure, such as a list of ports to which the file connects. This is in sharp contrast with standard datasets, where each sample has the same number of features, and each feature is a real number. We processed the data by a public implementation of hierarchical multi-instance learning (HMIL) \cite{pevny2017using}. Then we applied \textit{DeepTopPush} and \textit{Pat\&Mat-NP} at $\tau=10^{-3}$ and $\tau=10^{-2}$. The latter maximizes the true positives rate when the false positive rate is at most $\tau$. The minibatch size was $20000$, which allowed us to obtain precise threshold estimates and unbiased sampled gradients due to Section \ref{sec:bias1}.

Figure \ref{fig: roc curves} \textbf{B)} shows the performance on the test set. \textit{DeepTopPush} is again the best at low false positive rates. This is extremely important in cybersecurity as it prevents false alarms for malware. Even at the extremely low false positive rate $\tau=10^{-5}$, our algorithm correctly identified $46\%$ of malware. The circles denote the thresholds for which the methods were optimized. \textit{DeepTopPush} should have the best performance at the leftmost point, \textit{Pat\&Mat-NP} ($\tau=10^{-3}$) at $\tau=10^{-3}$ and similarly \textit{Pat\&Mat-NP}($\tau=10^{-2}$).

\subsection{Impact of enhancing the minibatch}\label{sec:delay}

The crucial aspect of \textit{DeepTopPush} is enhancing the minibatch by one sample. In all presented results with the exception of the Malware Detection, the minibatch contained only 32 samples. Then the discussion in Section \ref{sec:bias2} implies that \textit{Pat\&Mat-NP} equals to \textit{DeepTopPush} without enhancing the minibatch. In other words, \textit{Pat\&Mat-NP} uses Algorithm~\ref{alg1} while \textit{DeepTopPush} uses Algorithm \ref{alg2}. As Table~\ref{tab:Overall comparison} clearly shows that \textit{DeepTopPush} ourperforms \textit{Pat\&Mat-NP}, this implies that using the delayed values is beneficial.

Figure \ref{fig:thresholds2} shows explanation for this behaviour. The full blue line shows the behaviour of \textit{DeepTopPush} while the dotted grey line shows \textit{Pat\&Mat-NP}. As explained in the previous paragraph, their difference demonstrates the effect of enhancing the minibatch by one delayed value. The top subfigure compares thresholds with the true threshold (dashed black). While the threshold for \textit{Pat\&Mat-NP} jumps wildly, it is smooth for \textit{DeepTopPush}, and it often equals the true threshold. Theorem \ref{theorem:convergence} then implies that our sampled gradient is an unbiased estimate of the true gradient. This is even more pronounced in the bottom subfigure, which shows the angle between the true gradient and the computed gradient. This angle is important because \cite{nocedal2006numerical} showed that if this angle is uniformly in the interval $[0,90)$, then gradient descent schemes converge. This is precisely what happened for \textit{DeepTopPush}. When the threshold is correct, the true and estimated gradients are parallel to each other, and the gradient descent moves in the correct direction.

\begin{figure}[!ht]
    \centering
    \begin{tikzpicture}
        \begin{groupplot}[
            group style = {
                group size = 1 by 2,
                vertical sep = 5pt,
                x descriptions at = edge bottom,
                y descriptions at = edge left,
            },
            footnotesize,
            xtick={0,2,4, 6, 8, 10},
            width= 0.9\linewidth,
            tick pos = left,
            enlargelimits = false,
        ]

        \nextgroupplot[
            title = {},
            ylabel = {Threshold},
            ytick={0, 4, 8, 12},
            ymin=0,
            ymax=14.5,
            height=120,
        ]
            \addplot [lineC] table[x index=4, y index=9] {\thresholds};
            \addplot [lineA] table[x index=4, y index=10] {\thresholds};
            \addplot [lineB] table[x index=4, y index=11] {\thresholds};

        \nextgroupplot[
            title = {},
            xlabel = {Epoch},
            ylabel = {Gradient angle},
            ytick={0,45,90},
            ymin=-5,
            ymax=100,
            height=120,
        ]
            \addplot [lineC] table[x index=4, y index=3] {\thresholds};
            \addplot [lineA] table[x index=4, y index=2] {\thresholds};
            \addplot [lineB] table[x index=4, y index=12] {\thresholds};
            \addplot [lineB] table[x index=4, y index=13] {\thresholds};
        \end{groupplot}
    \end{tikzpicture}
    \caption{The thresholds (top) and angle between true and sampled gradients (bottom) for Algorithm \ref{alg1} (full blue) and Algorithm \ref{alg2} (dotted gray).}
    \label{fig:thresholds2}
\end{figure}

\section{Conclusions}

We proposed \textit{DeepTopPush} as an efficient method for solving the constrained non-decomposable problem of accuracy at the top, which focuses on the performance only above a threshold. We implicitly removed some optimization variables, created an unconstrained end-to-end network and used the stochastic gradient descent to train it. We modified the minibatch so that the sampled threshold (computed on a minibatch) is a good estimate of the true threshold (computed on all samples). We showed both theoretically and numerically that this procedure reduces the bias of the sampled gradient. The time increase over the standard method with no threshold is small. We demonstrated the usefulness of \textit{DeepTopPush} both on visual recognition datasets, a ranking problem and on a real-world application of malware detection.

\bibliographystyle{abbrv} 
\bibliography{references}

\appendix

\section{Source code}

To promote reproducibility, we provide one respository with the code
\begin{center}
  \url{https://github.com/VaclavMacha/AccuracyAtTop.jl}
\end{center}
and one repository with numerical experiments 
\begin{center}
  \url{https://github.com/VaclavMacha/AccuracyAtTopDeepTopPush_experiments.jl}
\end{center}

\section{Theorem \ref{theorem:convergence} for Rec@K}

The assumption of Theorem \ref{theorem:convergence} requires that the threshold is computing from negative samples and the objective for positive samples. This does not hold for Rec@K. We will show that we can obtain a similar result even for this case.

The proof of Theorem \ref{theorem:convergence} is based on Lemma \ref{lemma:convergence}. We will now obtain the variant of Lemma \ref{lemma:convergence} for Rec@K. First, we realize that if the threshold index $j^*$ corresponds to a negative sample, the computation will not change and therefore
$$
 \aligned
 &\EE\left(\nabla \hat L(\bm w) \mid \hat j=j^*\text{ is an index of a negative sample} \right) \\
 &\qquad=  \nabla L(\bm w).
 \endaligned
$$
On the other hand, when $j^*$ corresponds to a positive sample, it needs to be always present in the minibatch selection and there are effectively only $\nmin^+-1$ positive samples in the minibatch. Then
$$
 \aligned
 &\EE\left(\nabla \hat L(\bm w) \mid \hat j=j^*\text{ is an index of a positive sample} \right) \\
 &\qquad =  \frac{\nmin^+-1}{\nmin^+}\nabla L(\bm w).
 \endaligned
$$
Denote now $p$ the probability that the threshold corresponds to a positive sample. Then we have
$$
 \aligned
 \EE\left(\nabla \hat L(\bm w) \mid \hat j=j^*\right) &= (1-p)\nabla L(\bm w) + p\frac{\nmin^+-1}{\nmin^+}\nabla L(\bm w) \\
 &= \nabla L(\bm w) - \frac{p}{\nmin^+}\nabla L(\bm w).
 \endaligned
$$

Theorem \ref{theorem:convergence} will then be modified into
$$
 \aligned
 \bias(\bm w) &= \PP(\hat j\neq j^*) \left(\nabla L(\bm w) - \EE\left(\nabla \hat L(\bm w) \mid \hat j\neq j^*\right) \right) \\
 &\qquad - \PP(\hat j= j^*)\frac{p}{\nmin^+}\nabla L(\bm w).
 \endaligned
$$
We changed the result by adding the last term. Usually the training set contains much less positive than negative samples. This implies that $p$ is assumed to be small and the extra term is small as well. Thefore, this change should have a negligible impact on the theorem implications.

\section{Used network architecture}\label{app:network}

For 3A4, we preprocessed the input with $9491$ into a $100$-dimensional input by PCA. Then we used two dense layers of size $100\times 50$ and $50\times 25$ with batch-normalization after these layers. The last layer was dense.

For FashionMNIST, we used a network alternating two hidden convolutional layers with two max-pooling layers finished with a dense layer. The convolutional layers used kernels $5\times 5$ and had $20$ and $50$ channels, respectively. For CIFAR100 and SVHN2, we increased the number of hidden and max-pooling layers from two to three. The convolutional layers used kernels $3\times 3$ and had $64$, $128$, and $128$ channels, respectively. A more detailed description can be found in our codes online. We are fully aware that these architectures are suboptimal. Since the accuracy at the top needs to select only a few relevant samples and the rest of the dataset's performance is irrelevant, such a network can be used. Moreover, using a simpler network has the advantage of faster experiments.

For ImageNet, we merged all turtles into the positive class and all non-turtles into the negative class. Then we used the pre-trained EfficientNet B0, where we replaced the last dense layer with $1000$ outputs by a dense layer into a scalar output.

\section{Dataset summary}

Table \ref{tab:Datasets} summarizes the used datasets. The Malware Detection dataset was represented by JSONs, which contain varying number of features. Moreover, many features are not scalar but have some hierarchical structure as well.

\begin{table*}[!ht]
  \centering
  \begin{tabular}{@{}lcrrrrl@{}}
    \toprule
    && \multicolumn{2}{c}{Train} & \multicolumn{2}{c}{Test} & License\\
    \cmidrule(lr){3-4} \cmidrule(lr){5-6}
    \thead{Dataset} & \thead{$d$}
                    & \thead{$n$}
                    & \thead{$\frac{n^+}{n}$}
                    & \thead{$n$}
                    & \thead{$\frac{n^+}{n}$} 
                    & \\
    \midrule
    FashionMNIST & $28 \times 28 \times 1$
                  & 60 000
                  & 10.00\%
                  & 10 000
                  & 10.00\% 
                  & MIT \\
    CIFAR100 & $32 \times 32 \times 3$
              & 50 000
              & 1.00\%
              & 10 000
              & 1.00\% 
              & not specified \\
    SVHN2 extra & $32 \times 32 \times 3$
                & 604 388
                & 17.28\%
                & 26 032
                & 19.59\% 
                & not specified \\
    ImageNet & $62 720 \times 1$
              & 1 281 167
              & 0.51\%
              & 50 000
              & 0.50\% 
              & registration \\
    3A4 & $9491 \times 1$
        & 37 241
        & 0.98\%
        & 37 241
        & 1.07\% 
        & CC BY 4.0 \\
    Malware Detection & variable
                      & 6 580 166
                      & 87.22\%
                      & 800 346
                      & 91.80\% 
                      & proprietary\\
    \bottomrule
  \end{tabular}
  \caption{Summary of the used datasets with the number of features~$d$, number of samples~$n$ and the fraction of positive samples~$\frac{n^+}{n}$ in the training set.}
  \label{tab:Datasets}
\end{table*}

\end{document}